\documentclass{article} 
\usepackage{iclr2026_conference,times}

\usepackage{amsmath,amsfonts,bm}

\def\eqref#1{equation~\ref{#1}}

\def\1{\bm{1}}

\DeclareMathAlphabet{\mathsfit}{\encodingdefault}{\sfdefault}{m}{sl}
\SetMathAlphabet{\mathsfit}{bold}{\encodingdefault}{\sfdefault}{bx}{n}

\usepackage{url}
\usepackage{graphicx}
\usepackage{booktabs}
\usepackage{multirow}
\usepackage{array}
\usepackage{siunitx}
\usepackage{algorithm}
\usepackage{algpseudocode}
\usepackage{xcolor}
\usepackage{amssymb}
\usepackage{flafter}
\usepackage{placeins}
\usepackage[hidelinks]{hyperref}

\title{Rehearse: Stepping Back from the Confidence Cliff in Self-Improving Autoresearch}

\author{%
Ji Jiazhen \\
Tencent \\
\texttt{royji@tencent.com}
\And
Ding Shouhong \\
Tencent \\
\texttt{ericshding@tencent.com}
}

\iclrfinalcopy  
\begin{document}

\maketitle
\fancyhead[L]{}

\begin{abstract}
Autoresearch improves machine-learning code by proposing changes, running full
training jobs, and keeping changes that improve the metric. The efficiency of
this loop depends not only on generating ideas, but also on the agent's ability
to decide, before spending a training run, whether a proposed modification is
likely to work. We study how the reliability of this pre-execution judgment
changes over the course of an autoresearch trajectory.
In public AutoSOTA logs~\citep{autosota2026,autosota2026logs}, the fraction of helpful modifications falls from
70\% in the first two iterations to 43\% by iteration 6+. On 296 same-baseline
modification pairs from 39 paper-derived AutoSOTA tasks, each containing one
modification that improved the metric and one that did not, with measured
outcomes hidden, an LLM judge given
candidate rationales but no prior-attempt history reaches 79.5\%
accuracy on the pairs where strict consensus returns a verdict. On the full
366-pair benchmark, however, this ability weakens substantially late in the
loop. As successful changes accumulate, selective
accuracy---accuracy conditioned on a strict-consensus verdict---falls from
82.8\% to 56.9\%, while the judge remains
willing to decide. We call this operational pattern the \emph{confidence cliff}.
\textbf{Rehearse} implements the loop change as a lightweight skill for
autoresearch loops: propose several ideas, compare them before execution, run
the most promising, and judge with a focused memory of similar past attempts and
outcomes. This focused outcome memory raises late selective accuracy to
83.5\%. Across $4{,}000$ budgeted training runs over three loops, Rehearse
improves the endpoint under the same training-run budget on nanochat, image
classification, and time-series forecasting.
\end{abstract}

\section{Introduction}
\label{sec:intro}

Autoresearch lets an LLM agent improve a machine-learning codebase against a
measurable objective: propose a modification, run a full training job, and keep it
if the metric improves. We study this setting rather than systems whose output is
a novel idea or a paper \citep{aiscientist2024, evoscientist2026}. Each unsuccessful
modification consumes a training run before its patch is reverted.

In the public AutoSOTA loops, wasted training runs become more common late. The
logs~\citep{autosota2026,autosota2026logs} show that the fraction of helpful
modifications falls from 70\% in iterations 1--2 to 43\% by iteration 6+,
and their mean gain shrinks from 3.6\% to 0.3\% (Figure~\ref{fig:figure0},
Appendix~\ref{sec:appendix-decay}). Late in these runs, most training jobs
produce changes that are subsequently reverted, and the few that work pay less
than early changes. The question is
not only how to generate more ideas, but how to decide which idea deserves a
training run. We therefore study pre-execution judgment: the decision a
compute-bound loop makes before spending a training-run budget. In this setting, success rate
is coupled to the agent's pre-run discrimination ability: whether it can tell
which proposed experiment is likely to work.

Recent work has established that pairwise preferences can filter candidate ML
solutions before execution~\citep{zheng2026foreagent}. We ask a complementary
question: whether this judgment remains reliable as a loop accepts changes and
moves through an increasingly history-dependent optimization trajectory. The
AutoSOTA logs let us test this depth dependence directly~\citep{autosota2026,autosota2026logs}. We build an outcome-labeled
benchmark from 39 paper-derived AutoSOTA tasks, centered on 296 strictly
same-baseline pairs, each containing one modification that improved the metric
and one that did not, with an auxiliary 70-pair
worked-versus-worked ranking probe. With an outcome-blind candidate rationale but
no memory, a judge comparing candidates before execution selects much better than
random. For an illustrative 10-candidate, keep-three deployment setting,
Appendix~\ref{sec:appendix-precision} reports a coverage-aware tournament
calibration from pairwise selective accuracy to Precision@3. In the live
ablation below, explicit
selection also beats taking the first proposal. The judgment is therefore
actionable: propose several candidates, compare them, and spend training runs on
the few that look most promising.

\begin{figure}[!htbp]
\centering
\includegraphics[width=0.75\linewidth]{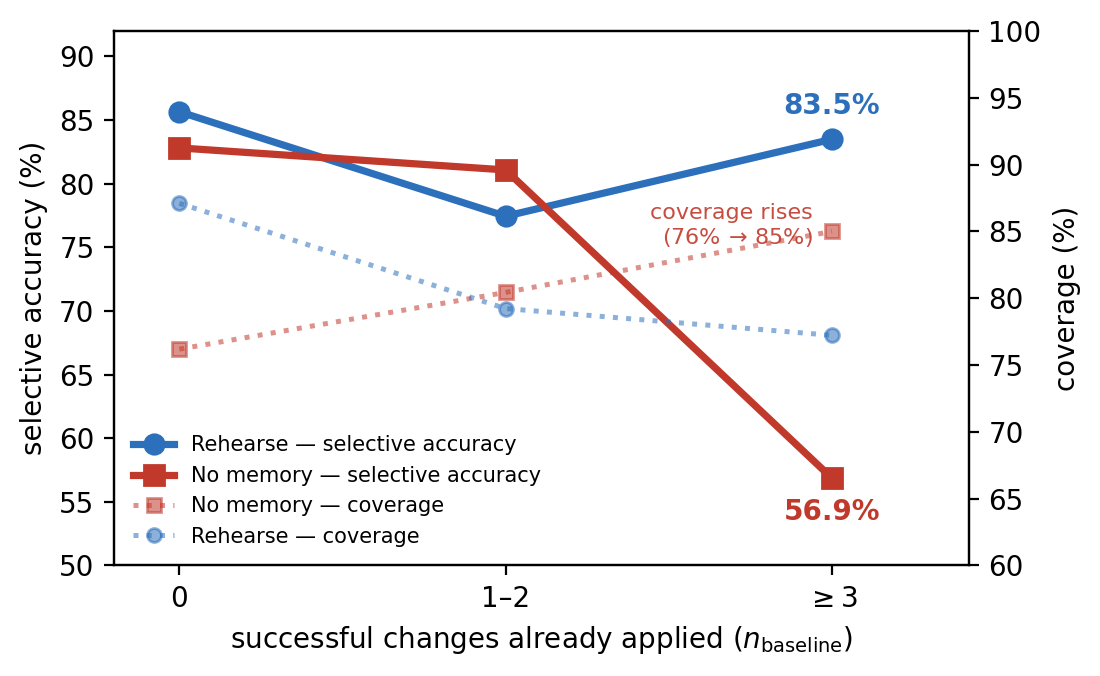}
\caption{\textbf{Pre-execution judgment fails late.} Selective accuracy of the pre-run
judgment on the 366-pair benchmark, by the number of successful changes already
accumulated ($n_{\mathrm{baseline}}$). We use \emph{confidence cliff} for the regime
where selective accuracy falls while coverage does not. Without memory,
selective accuracy collapses $25.9$ points once three or more successes have
accumulated, while coverage (dotted, right axis) rises from $76\%$ to $85\%$;
the focused store holds selective accuracy at $83.5\%$ and lowers coverage
instead.}
\label{fig:cliff}
\end{figure}

This judgment degrades at greater loop depth. Later candidates build on more
successful changes and must be judged against a longer history. On the same 366-pair
proxy, a memoryless judge has 82.8\% selective accuracy at cold start but only
56.9\% once three or more successes have accumulated, while committing just
as readily (Figure~\ref{fig:cliff}). It becomes more willing to return a
verdict, even as those verdicts become less reliable. We call this operational
pattern the \emph{confidence cliff}. Late-stage waste therefore has at least two
contributors: useful
changes become rarer, and the agent becomes less reliable at judging which
remaining changes are worth a run. Rehearse targets the judgment component: as
gains become sparse, it spends runs more selectively on the candidates that
remain.

On the same late decisions, focused outcome memory restores selective accuracy
while less focused history forms lag. This comparison motivates presenting
candidate-relevant modifications and their outcomes instead of the entire trace.
Rehearse combines an explicit pre-run comparison step with a judge that sees only
a focused set of relevant past outcomes. The agent
proposes several modifications, compares them before execution, and runs only the
most promising. Each executed modification is then recorded as what changed and
whether it worked, and the next judgment sees only the few most similar records.

\paragraph{Contributions.}
\textbf{(1)~We benchmark pre-execution judgment across autoresearch depth.} We
construct an outcome-labeled benchmark of incremental code modifications and
measure the same judgment as accepted changes accumulate, rather than only at a
single search state.
\textbf{(2)~This judgment falls off a confidence cliff.} On the AutoSOTA
outcome benchmark (39 paper-derived tasks, three judge models), the same
judgment keeps committing as history accumulates while its selective accuracy falls from
82.8\% to 56.9\%.
\textbf{(3)~Focused outcome memory preserves the judgment.} On the single-turn
benchmark, focused retrieval gives the highest cliff selective accuracy among the tested
history forms (83.5\%, versus 80.9\% for the closest summary baseline).
\textbf{(4)~We implement and validate Rehearse as an autoresearch skill and will
release it with the benchmark.} The skill wraps proposal, pre-run selection, and focused outcome
memory. Across $4{,}000$ budgeted training runs over three loops, Rehearse
improves the endpoint under the same training-run budget on nanochat, image
classification, and time-series forecasting.

\section{Related Work}
\label{sec:related}

\subsection{Measurable-objective autoresearch and pre-execution selection}
\label{sec:related-measurable}

We study autoresearch loops that optimize a \emph{measurable} objective: a
validation loss or accuracy returned by a full training run. Karpathy's
nanochat loop edits \texttt{train.py}, runs one training job, reads back
\texttt{val\_bpb}, and keeps the edit only if the score
improves~\citep{karpathy2026autoresearch}. AIDE casts ML engineering as search
over a scored solution tree, drafting, debugging, or improving code from selected
nodes~\citep{aide2025}. The
evolutionary line (FunSearch~\citep{funsearch2024}, AlphaEvolve~\citep{alphaevolve2025},
and the open CodeEvolve~\citep{codeevolve2025}) keeps archives of scored
programs and uses LLM generation, mutation, or crossover to propose variants.
Aster~\citep{aster2026} and Bilevel
Autoresearch~\citep{bilevelautoresearch2026} push on iteration efficiency, the
latter on the same nanochat benchmark we use. MLR-Copilot~\citep{mlrcopilot2024}
plans ideas before running them but grounds candidates with external research
artifacts such as prototype code, models, and datasets.
ML-engineering benchmarks such as MLE-Bench~\citep{mlebench2025} evaluate agents
that iterate on machine-learning experiments, and Bayesian-optimization or
AutoML systems, including LLM-augmented variants such as
LLAMBO~\citep{llambo}, also choose experiments before paying for evaluation.
Those systems usually operate over numerical or structured configuration spaces.

ForeAgent is the closest prior system: it compares fully instantiated ML solution
code before execution and uses confidence-gated pairwise filtering to decide
which candidates to verify in an AIDE-style search~\citep{zheng2026foreagent}.
Its offline and online results establish that solution-level preference prediction
can improve search efficiency. Rehearse targets an earlier, state-matched
decision. ForeAgent withholds measured outcomes, but its offline corpus
exhaustively recombines curated complete workflows within each task; the judge
sees two solution codes and a verified data report. Rehearse instead compares
pending incremental proposals---descriptions, hypotheses, and implementation
plans---before either patch is applied, with each primary pair sharing the exact
same accepted baseline. It therefore asks which next modification is worth a
run, rather than which completed solution is better. This construction also
reveals how judge accuracy and coverage change as accepted improvements
accumulate, whereas ForeAgent's online evaluation reports search outcomes rather
than depth-indexed judge behavior. ForeAgent establishes solution-level
feasibility; Rehearse studies proposal-level reliability across evolving states
and tests candidate-conditioned retrieval of the same loop's prior outcomes as an
intervention for late-stage degradation.

Across these systems, stored evidence is usually an evaluated artifact or a
summary rather than a focused pre-run outcome memory. AIDE stores a scored
solution tree and summaries, evolutionary systems keep scored-program archives,
and a greedy loop carries the trace of accepted or reverted edits. Their search
decisions are also mostly driven by outcomes already measured: a tree expands
after evaluation, evolution samples from evaluated archives, and a greedy loop
keeps an edit only after training. MLR-Copilot is a useful exception because it
plans before execution, but its retrieval grounds candidates in external evidence
rather than the loop's own outcome history~\citep{aide2025,funsearch2024,alphaevolve2025,codeevolve2025,mlrcopilot2024}.
These systems therefore provide memory and selection, but those mechanisms alone
do not characterize whether pre-execution judgment remains reliable as the
accepted state evolves. Rehearse makes this depth-indexed reliability curve the
object of study and uses focused outcome retrieval as the intervention.

\subsection{Autoresearch that writes papers}
\label{sec:related-papers}

A second line automates broader scientific discovery and research write-up. The
AI Scientist~\citep{aiscientist2024} and its successor v2~\citep{aiscientistv2},
AI-Researcher~\citep{airesearcher2025}, and
AutoResearchClaw~\citep{autoresearchclaw2026} generate ideas, run experiments,
and draft or evaluate papers. The AI Co-Scientist~\citep{aicoscientist2026}
instead generates, critiques, ranks, and evolves research hypotheses and
proposals for scientist-guided validation. These systems primarily evaluate
end-to-end research, paper, or hypothesis quality, rather than the reliability of
pre-execution accept/reject decisions under a fixed measurable training metric.
Persistent-memory research systems such as EvoScientist~\citep{evoscientist2026}
and AutoResearchClaw~\citep{autoresearchclaw2026} distill past ideas,
experiments, or mistakes to improve later research runs. Rehearse is narrower:
it records concrete candidate-change and outcome pairs, then reads those records
before paying for a specific ML training run.

\subsection{Memory and reflection in agents}
\label{sec:related-memory}

Our store builds on memory, reflection, and test-time refinement work.
Generative Agents store and retrieve natural-language experience for believable
behavior~\citep{generativeagents2023}; MemGPT pages context
in and out like an operating system~\citep{memgpt}; Voyager keeps a library of
executable skills~\citep{voyager}; Reflexion reflects on task feedback across
trials~\citep{reflexion}; and Self-Refine performs test-time self-feedback and
refinement~\citep{selfrefine2023}. Devil's Advocate moves reflection before an
action by anticipating failures and remedies~\citep{devilsadvocate2024}. We
borrow the idea of reusing prior experience, but the decision is different.
These methods do not store measured code-change outcomes in order to gate a
pending training run. In autoresearch the waste is the execution itself, so
Rehearse reads memory \emph{before} executing.

Recent work stores typed records of agent experience as well.
Trajectory-Informed Memory~\citep{trajectory-memory2026} extracts strategy,
recovery, and optimization learnings from execution trajectories and retrieves
them on app-and-API tasks. Concurrent EvolveMem uses an autoresearch loop to
adapt the retrieval architecture itself~\citep{evolvemem2026}. Rehearse instead
keeps retrieval fixed and studies a different decision---whether a proposed
code modification is worth the compute to run at all---on a measurable ML-code
metric.

\subsection{Self-improving loops and self-modifying agents}
\label{sec:related-selfimprove}

Our setting is itself a self-improving loop: each iteration the agent improves
the code it will run next~\citep{karpathy2026autoresearch}. A stronger line takes
self-improvement down to the agent itself---fine-tuning on self-generated
rationales or self-judged reward data~\citep{star2022,selfrewardingLM2024}, or
rewriting the agent's own source code~\citep{dgm2025}. Rehearse studies
sample-efficient self-improvement by changing the loop's selection and memory
while holding the agent weights and source code fixed. Within this setting, we
study the confidence cliff:
as a loop accumulates its own successes, its judgment of what to try next becomes
more committal and less reliable.

\section{Rehearse}
\label{sec:method}

\begin{figure}[t]
\centering
\includegraphics[width=0.7\linewidth,height=7cm]{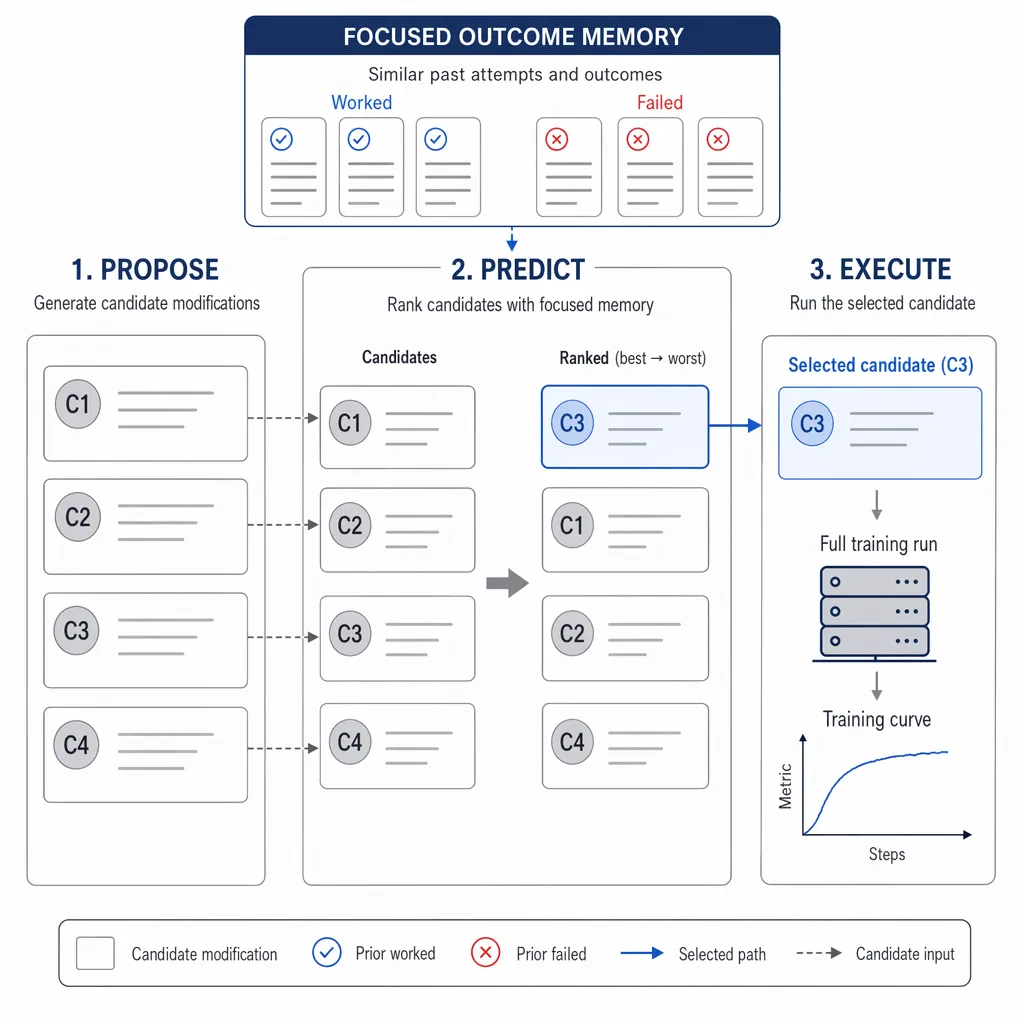}
\caption{\textbf{Overview of Rehearse.}
Rehearse turns a one-idea-then-run cycle into Propose--Predict--Execute:
propose multiple candidate modifications, rank them with focused outcome memory,
run the top candidate, and write its result back to the store.}
\label{fig:overview}
\end{figure}

A vanilla autoresearch loop proposes a modification, runs a full training job,
and keeps the change if the metric improves. Rehearse adds one operation and one
state object. The operation is \emph{predict}: before execution, a judge ranks
candidate modifications and sends only the top $k$ to training (all live
experiments here use $k=1$). The state object is a focused outcome store: it
records executed attempts and retrieves similar outcomes as precedent. We call
the resulting loop \emph{Propose--Predict--Execute}
(Algorithm~\ref{alg:ppe}, Appendix~\ref{sec:appendix}).

\subsection{Predicting before executing}
\label{sec:method-predict}

The predict step externalizes the comparison that a one-idea loop leaves
implicit. Running a candidate consumes a training run, whereas ranking uses
model inference. An exhaustive strict-consensus tournament over $N$ candidates
uses $2\binom{N}{2}$ judge calls---$20$ when $N=5$, as in our live loops.
Rehearse ranks the candidates first and executes only the top $k$.

We implement the ranking as a pairwise tournament: a judge LLM is shown the
shared task context---target paper/codebase summary, target metric, and baseline
state---and two candidate modifications, and asked which improves the metric more.
Each candidate carries a description, a \emph{hypothesis} (the mechanistic reason
the change should help), an implementation field, and a coarse type. This
rationale channel lets the judge assess the proposed mechanism rather than only
the edit. Removing the hypothesis and implementation produces the largest drop
among prompt-component ablations (Section~\ref{sec:single}). To mitigate position
bias, each unordered pair is
judged in both presentation orders under \emph{strict consensus}. An agreed
verdict gives one vote to the winner; disagreement gives no vote to either
candidate. Candidates are ranked by vote count and the top $k$ run. We detail and
validate this protocol in Section~\ref{sec:single-protocol}. Pairwise auditing is
important here because global judge agreement need not predict within-pool
best-of-$N$ selection quality~\citep{landesberg2026judge}.

The store matters later in the run, when focused precedents
change pairwise votes or down-rank candidates that resemble exhausted directions.

\subsection{The focused store}
\label{sec:method-store}

The autoresearch loops we study usually carry history as a raw trace---past
changes and scores appended to the prompt or log~\citep{autosota2026,autosota2026logs,aide2025,karpathy2026autoresearch}.
Rehearse keeps a store but reads it selectively. When a new candidate is judged,
retrieval is restricted to attempts from the same task that were executed before
the current step; the store returns only past attempts within cosine similarity $0.40$
(Appendix~\ref{sec:appendix-threshold}). Each retrieved attempt is serialized as
one line containing the change and its binary outcome, so the judge receives a
small set of candidate-relevant precedents rather than a full history. Adding
reasons, summaries, or the full applied-change list lowers cliff selective
accuracy (Section~\ref{sec:single-cliff}, Appendix~\ref{sec:appendix-why}).

\subsection{Delivery}
\label{sec:method-delivery}

Rehearse is delivered as a skill, not a model rewrite: no fine-tuning
and no change to the target training code. The skill wraps the controller around
an existing autoresearch loop, replacing its decision policy between proposal and
execution and adding a focused outcome store backed by a few plain files. We
isolate the pieces on a controlled benchmark (Section~\ref{sec:single}) and run
the full skill against the vanilla loop on real tasks (Section~\ref{sec:multi}).

\section{Single-turn evaluation}
\label{sec:single}

To study the predict step in isolation, we strip the loop to one decision: given
two candidate modifications, which is better to run? The benchmark lets us measure
that judgment against ground truth, turn each part of Rehearse on or off, and ask
three questions: whether pre-execution judgment is useful when made explicit,
where it breaks as history accumulates, and which memory form keeps it usable.

\subsection{A benchmark of real modification outcomes}
\label{sec:single-benchmark}

We build the benchmark from public AutoSOTA logs~\citep{autosota2026,autosota2026logs}, where each
task starts from the repository associated with a SOTA paper and asks the
optimizer to improve its measured ML metric. We treat the task/codebase, not the
paper text, as the unit of analysis. Every proposed code modification has a
measured outcome. The changes span optimizer, schedule, architecture,
regularization, augmentation, loss, precision, and batching rather than one narrow
family, and each is proposed on top of an already strong baseline, so whether it
helps is genuinely unknown in advance.

We label a modification \emph{worked} iff it strictly improves the task metric
over the current accepted baseline. Crashes, timeouts, invalid patches, and
non-improvements are labeled \emph{did not work}; we use these labels throughout.
We use two pair types. The main set is \emph{type-1}: one modification worked and
one did not work. We form these pairs only when the two candidates are evaluated from
the same accepted baseline state, so both sides share the same task, codebase,
metric, starting point, and remaining headroom. Attempts labeled did not work do
not update that baseline, so such an attempt before the next successful change can still be
paired with that later success without changing the state being judged. This
strict same-baseline construction yields 296 pairs containing one worked
modification and one that did not work.

We also include a smaller auxiliary \emph{type-2} probe: both modifications
worked, and the one with the larger measured improvement is labeled correct.
Because successful modifications update the accepted state, a type-2 pair can
compare two successful attempts from successive accepted states, where the earlier
success may already be part of the later state. We restrict type-2 to cases where
that earlier success is only a small incremental gain, keeping the later state
close enough for an auxiliary ranking probe. The main filtering claim rests on
type-1. For every pair, the judgment-time history is cut off before the earlier
of the two candidate evaluations, and both candidates in the pair are excluded
from retrieval. Together this yields 366 pairs from 39 paper-derived AutoSOTA
tasks (296 type-1, 70 type-2).

Candidates are shown as pending proposals with description, hypothesis,
implementation, and type. The hypothesis field is generated after log collection
by a separate LLM that sees the candidate, task context, implementation, and
shared baseline state, but is blinded to the measured outcome, metric delta,
failure reason, and whether the candidate was kept. Measured result and failure
reason are stripped. The judge never sees an outcome.

The corpus comes from an LLM-driven optimizer, so the benchmark measures how
well an agent predicts the outcome of \emph{LLM-proposed} modifications from
outcome-blind candidate rationales, approximating the decision Rehearse makes
before execution. In the reported benchmark, each judged candidate is shown the
task/codebase context and its outcome-blind rationale. The no-memory condition in
Table~\ref{tab:baselines} is therefore the baseline for a well-informed pre-run
judge without prior attempts; the memory configurations add history on top of the
same candidate information.

\subsection{Strict-consensus judging}
\label{sec:single-protocol}

LLM judges are known to favor whichever option is presented first
\citep{zheng2023judging, wang2024position}. We mitigate this with
\emph{strict consensus}: each pair is judged twice, in both orders, and the
verdict counts only when the two orders agree; otherwise the judge abstains.
Following selective-prediction terminology~\citep{geifman2017selective}, we call
the fraction of pairs receiving a verdict \emph{coverage} and the accuracy
conditioned on those verdicts \emph{selective accuracy}. Every single-turn
benchmark number reported here is averaged over three different judge models,
so it does not rest on one model's idiosyncrasies. Strict consensus also stabilizes the measurement: it cuts the
run-to-run spread from 5.1 to 1.7 points, at the cost of abstaining on the pairs
that are sensitive to presentation order.

\begin{table}[!t]
\centering
\small
\setlength{\tabcolsep}{4.5pt}
\caption{\textbf{How history is shown matters in the late regime.} Pre-run judge
performance on the full benchmark and the deep/cliff bucket
($n_{\mathrm{baseline}}\!\ge\!3$). Coverage is the verdict rate under strict
consensus; selective accuracy is accuracy conditioned on a verdict.}
\label{tab:baselines}
\begin{tabular*}{0.94\linewidth}{@{\extracolsep{\fill}}l
S[table-format=2.1]
S[table-format=2.0]
S[table-format=2.1]@{}}
\toprule
Memory shown at decision & {Overall sel. acc. (\%)} & {Cliff cov. (\%)} & {Cliff sel. acc. (\%)} \\
\midrule
No memory (full rationale, no history)  & 77.6 & 85 & 56.9 \\
Full-history dump (AIDE-inspired)     & 82.0 & 80 & 70.8 \\
Self-reflection buffer (Reflexion-inspired) & 81.0 & 90 & 74.1 \\
LLM summary of history                & 81.9 & 78 & 80.9 \\
\textbf{Rehearse (focused retrieval, ours)} & \bfseries 82.3 & 77 & \bfseries 83.5 \\
\bottomrule
\end{tabular*}
\end{table}

\subsection{Pre-execution judgment is useful when made explicit}
\label{sec:single-selection}

Before seeing any execution result, a judge with candidate rationales but no
history distinguishes worked modifications from those that did not work at 79.5\% selective accuracy;
Rehearse raises this to 84.2\% (Section~\ref{sec:single-type2}). The benchmark decision is
pairwise, so Appendix~\ref{sec:appendix-precision} uses a coverage-aware
tournament model to calibrate the same signal to
higher expected success@1 in the live-sized 5-candidate setting and higher
expected Precision@3 in an illustrative 10-candidate, keep-three pool. We use
this calculation only to make the pairwise metric concrete;
Section~\ref{sec:multi} measures the live-loop effect when candidates are
generated online.

\subsection{The confidence cliff}
\label{sec:single-cliff}

Pairwise selective accuracy measures how often the judge selects the better of
two modifications. Every configuration in Table~\ref{tab:baselines} receives the
candidate \emph{rationale} (hypothesis and implementation); the table varies only
the history supplied at judgment time. With no history the judge commits to 85\%
of deep calls and reaches
56.9\% selective accuracy. An author-constructed full-history dump inspired by
AIDE's scored-tree summaries~\citep{aide2025} reaches 70.8\%; a self-reflection
buffer inspired by Reflexion~\citep{reflexion} reaches 74.1\% but commits most
eagerly, on 90\% of calls; and an LLM summary reaches 80.9\%. These baselines
adapt the cited systems' history representations to our common judge prompt;
they are not full reproductions of those systems. Rehearse retrieves only the few prior
attempts similar to each candidate and returns them as raw change-and-outcome
text, reaching 83.5\% selective accuracy while committing on the fewest deep calls. Overall averages
hide most of the late-stage failure; the cliff bucket is where the strategies
separate.

Performance does not increase monotonically with the amount of memory supplied.
The full-dump and Rehearse conditions draw from the same underlying outcome
store, but Rehearse presents only candidate-relevant records to the judge. Adding structure to
the dump does not recover it: appending similarity-retrieved failure reasons to
the full history reaches only 73.3\% on the cliff. The summary baseline comes
closest, but fixed-length summaries can omit the relevant precedent as history
grows and require a generation call each step. Focused retrieval returns that
narrow slice directly, regardless of history length.

We bucket each decision by $n_{\mathrm{baseline}}$, the number of successful
changes already in place when it is made (Figure~\ref{fig:cliff}). Without memory,
selective accuracy holds near 82.8\% while little has accumulated, then falls to 56.9\% once
three or more successes are in place. Its willingness to commit moves the other
way: as selective accuracy collapses, the rate at which the judge returns a verdict under
strict consensus rises from 76\% on shallow decisions to 85\% on deep ones. It
grows more committal as it grows less reliable. We call this the
\emph{confidence cliff}. The gain is not only abstention: on the deep calls both
judges commit to, accuracy rises from 63.4\% to 82.5\%, while coverage changes
from 85\% to 77\%. The few calls Rehearse declines are ones the memoryless judge
mostly gets wrong, so abstention adds to the gain rather than explaining it. A
similar pattern appears qualitatively in live loops, where a memoryless agent
re-runs directions that previously produced did-not-work attempts
(Section~\ref{sec:multi-case}).

Table~\ref{tab:ablation} reports leave-one-out ablations of Rehearse's components.
Removing the candidate rationale, retrieved text, or focused retrieval
substantially lowers selective accuracy.
Removing individual outcome cues lowers selective accuracy less sharply and,
when did-not-work precedents are removed, raises coverage while selective accuracy
falls. Adding failure explanations
also hurts: the recorded failure reason lowers cliff selective accuracy from
83.5\% to 80.6\%, and to 54.8\% on
calls that carry such a reason. Rehearse therefore keeps only the change and
outcome. These comparisons attribute the gain to focused similar changes and
outcomes rather than memory volume alone.
Appendix~\ref{sec:appendix-why} analyzes the components.

\subsection{Classification and ranking behavior}
\label{sec:single-type2}

The benchmark's two pair types show what the judge is doing. On \emph{type-1}
pairs, where one modification worked and the other did not, selective accuracy rises from
79.5\% without the store to 84.2\% with it. On \emph{type-2} pairs, where both
worked and the judge must pick the larger gain, selective accuracy is lower but still
moves with the store, from 67.4\% to 74.1\%. Because type-2 is an auxiliary
ranking probe, the cleanest pre-execution signal is classification:
separating modifications likely to work from those unlikely to work before any run.
That is the late-run selection problem: filtering the many unlikely candidates
recovers more compute than precisely ranking the few ideas that work.

\section{Multi-turn evaluation}
\label{sec:multi}

Section~\ref{sec:single} shows that pre-execution judgment is useful, degrades
later in the loop, and improves with focused outcome memory. The single-turn
benchmark measures this judgment directly. The live loops evaluate the full
path-dependent search process, which also depends on proposal quality and
execution.

\begin{figure}[t]
\centering
\includegraphics[width=\linewidth]{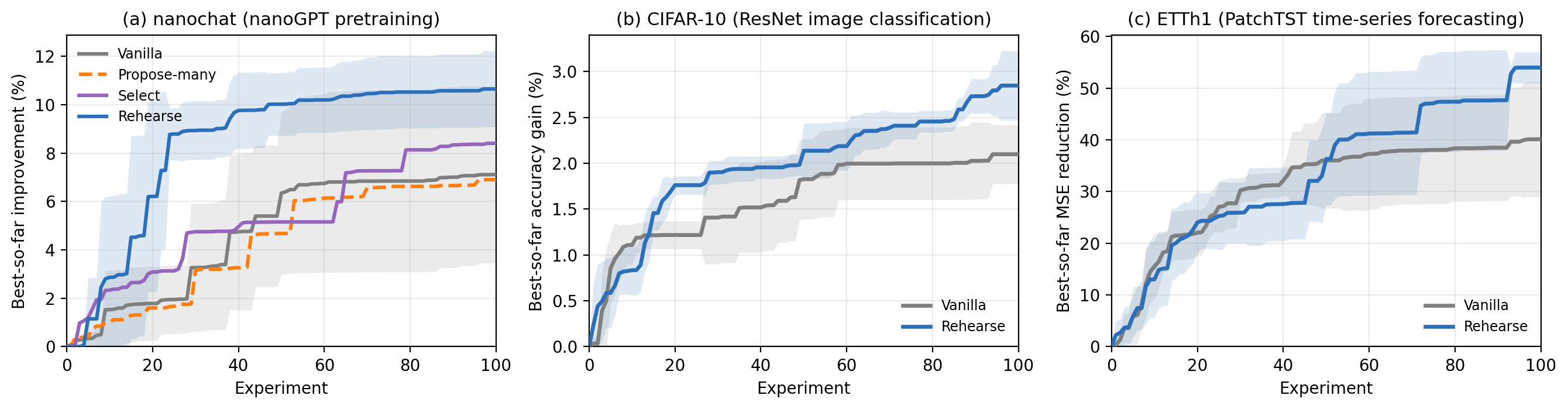}
\caption{\textbf{Pre-run selection improves live-loop search efficiency.}
Best-so-far improvement over baseline in live loops, mean over five seeds
($\pm$ one std). \textbf{(a)}~On nanochat, propose-many tracks vanilla, selection
front-loads progress, and the focused store (Rehearse) achieves the highest
endpoint. \textbf{(b--c)}~On two additional loops (CIFAR-10, ETTh1),
Rehearse reaches a better endpoint under the same 100-experiment training-run
budget, with lower observed seed-to-seed spread on ETTh1.}
\label{fig:live}
\end{figure}

\subsection{Setup}
\label{sec:multi-setup}

We run Karpathy's nanochat pretraining loop~\citep{karpathy2026autoresearch}, scored by validation
bits-per-byte (\texttt{val\_bpb}, lower is better) under a fixed
$30$-minute-per-experiment budget, for 100 experiments. To separate Rehearse's
two moves we run four configurations that share one idea-generating model and
differ only in how the step to run is chosen: \emph{vanilla} proposes one idea
and runs it; \emph{propose-many} proposes several and runs the first;
\emph{select} proposes several and runs the predict step's pick; \emph{Rehearse}
adds the store on top of select. We report five seeds each, as mean
relative improvement over each run's own baseline. We then compare \emph{vanilla}
and \emph{Rehearse} on two further loops in Section~\ref{sec:multi-general}. In
total, the multi-turn evaluation runs
$4{,}000$ budgeted training runs: $4$ nanochat configurations $\times$ $5$ seeds,
plus $2$ configurations $\times$ $5$ seeds on each of the two additional loops, all
for $100$ experiments.

\subsection{End-to-end search efficiency}
\label{sec:multi-efficiency}

Proposing several ideas and running the first, without selecting, ends close to
vanilla ($6.9\%$ vs $7.1\%$ mean improvement). The predict step makes
the larger pool useful: selecting rather than taking the first lifts the final
improvement to $8.4\%$. The store adds further improvement, to
$10.7\%$ ($+2.3$ pp over select), with the store ahead on every seed.

Selection and memory affect different parts of the trajectory. Selection
accelerates early search, while the store contributes additional gains later in
the run. On nanochat, Rehearse has cut
\texttt{val\_bpb} by 8.9\% at experiment~30, against vanilla's 3.3\%. Much of this
early gap comes from a batch-size reduction---the largest single change in these
runs---that the wider candidate pool surfaces earlier. This pattern is specific
to nanochat: where no single change dominates, as on the two loops below, the
advantage accrues later. Past experiment~fifty, vanilla largely plateaus while
Rehearse continues to improve.

\subsection{Generalization to two additional testbeds}
\label{sec:multi-general}

To check that the result is not tuned to nanochat or its model, we repeat the
endpoint comparison on two unrelated loops: CIFAR-10 classification~\citep{krizhevsky2009learning}
with \texttt{glm-5.1}~\citep{zai2026glm51docs} and ETTh1 forecasting~\citep{zhou2021informer}
with the CodeBuddy alias \texttt{deepseek-v4}~\citep{deepseek2026modeldocs,deepseekai2026deepseekv4}, versus
\texttt{hy3-preview}~\citep{tencent2026hy3preview} on nanochat.\footnote{These
names are serving endpoints or aliases rather than frozen checkpoints. The
citations identify exact public documentation where available and otherwise the
documented model family; Appendix~\ref{sec:appendix-repro} states this distinction.}
Each runs 100 experiments, comparing
\emph{vanilla} against \emph{Rehearse} over five seeds (Figure~\ref{fig:live}b--c).

Rehearse is higher on both. On CIFAR-10 it improves accuracy by $2.85\%$ over baseline
against vanilla's $2.10\%$; on ETTh1 it cuts MSE by $54.0\%$ against $40.1\%$.
Unlike nanochat, neither loop has a single dominant early change: the curves track
each other through the first half and separate later. After experiment~50,
Rehearse realizes about $2$--$4.3\times$ vanilla's remaining gain ($+0.7$ vs $+0.3$ points
on CIFAR, $+17.7$ vs $+4.1$ on ETTh1), in a regime analogous to the sparse late
stage identified by the single-turn benchmark. Rehearse reaches vanilla's final level after $54$
experiments on CIFAR and $63$ on ETTh1, compared with $100$ for vanilla ($46\%$
and $37\%$ fewer experiments). Per seed
(Table~\ref{tab:general}, Appendix~\ref{sec:appendix-perseed}), Rehearse ends
ahead on $9$ of $10$ additional-loop runs. The exception is a high-variance vanilla ETTh1
seed ($59.9\%$) while its other four land between $31$ and $45\%$; Rehearse is
tighter in this five-seed sample ($\pm3.0$ vs $\pm11.2$). This is consistent
with the store improving the endpoint and lowering observed seed-to-seed spread,
but the reliability claim should be read with the seed-level limitation in
Section~\ref{sec:discussion}.

\begin{table}[H]
\centering
\small
\setlength{\tabcolsep}{5pt}
\caption{\textbf{Focused outcome memory keeps late judgment reliable.}
Cliff-bucket ablation ($n_{\mathrm{baseline}}\!\ge\!3$). Coverage is the verdict
rate under strict consensus; selective accuracy is accuracy conditioned on a verdict.}
\label{tab:ablation}
\begin{tabular*}{0.94\linewidth}{@{\extracolsep{\fill}}>{\raggedright\arraybackslash}p{0.57\linewidth}
S[table-format=2.0]
S[table-format=2.1]@{}}
\toprule
Configuration & {Coverage (\%)} & {Selective acc. (\%)} \\
\midrule
\textbf{Rehearse (full)} & 77 & \bfseries 83.5 \\
\addlinespace[2pt]
\multicolumn{3}{@{}l}{\emph{Remove what the judgment reasons from}} \\
\quad $-$ candidate rationale                  & 78 & 70.2 \\
\quad $-$ idea text ($\to$ similar-count only) & 80 & 70.8 \\
\quad $-$ focused retrieval ($\to$ full dump)  & 80 & 70.8 \\
\addlinespace[2pt]
\multicolumn{3}{@{}l}{\emph{Remove outcome information}} \\
\quad $-$ outcome label                        & 77 & 77.5 \\
\quad $-$ did-not-work precedents (worked only) & 82 & 78.9 \\
\quad $-$ worked precedents (did-not-work only) & 74 & 76.1 \\
\quad $-$ extra neighbors (top-1 only)         & 77 & 80.5 \\
\addlinespace[2pt]
\multicolumn{3}{@{}l}{\emph{Add explanatory content}} \\
\quad $+$ recorded failure reason             & 77 & 80.6 \\
\quad $+$ curated failure reason              & 81 & 78.7 \\
\midrule
No history (full rationale)                  & 85 & 56.9 \\
\bottomrule
\end{tabular*}
\end{table}

\subsection{Focused memory enables conditional reuse}
\label{sec:multi-case}

Focused memory makes candidate-level decisions rather than blacklisting entire
directions. In a representative vanilla run, the loop repeatedly retuned
hyperparameter families that had produced did-not-work attempts; Rehearse did so
far less. A direction-level blacklist
would nevertheless remove useful candidates: lowering weight decay (0.2 to 0.15)
was reverted, while raising it was kept across three steps (0.2 to 0.25 to 0.30
to 0.35), and $87\%$ of Rehearse's kept changes belonged to families that had
also produced did-not-work attempts. This supports candidate-level outcome
judgment (Section~\ref{sec:single-type2}) over direction-level deduplication.

\section{Limitations}
\label{sec:discussion}

\paragraph{Evidence and scope.} Live aggregates average five seeds per
configuration, with endpoint effects, standard deviations, and per-seed outcomes
reported. The single-turn benchmark isolates pre-execution judgment; live loops
also reflect proposal quality, path dependence, implementation errors, and metric
noise. The evidence covers three loops with measurable outcomes;
novelty- or paper-scored systems and larger-scale seed studies remain outside its scope.

\paragraph{Candidate pools and cost.} Rehearse selects only from proposer
candidates. The nanochat ablation separates breadth, selection, and memory, but
other memory forms and candidate counts are untested live. Reported budgets count
training runs, excluding proposal and judge inference
(Appendix~\ref{sec:appendix-repro}).

\paragraph{Cliff measurement.} The depth variable $n_{\mathrm{baseline}}$ is
observational, so later decisions may differ in headroom, modification family,
and difficulty. Primary type-1 pairs share a baseline and use outcome-blind
rationales, history cutoff, and length balancing; all memory variants face the
same decisions, and type-2 pairs are reported separately. The benchmark is thus
a controlled proxy at run depth, not a randomized causal estimate; live loops
test downstream utility.

\section{Conclusion}
\label{sec:conclusion}

Autoresearch efficiency depends on pre-run selection as well as idea generation.
We measure this judgment across loop depth and identify a late regime where
selective accuracy falls while coverage stays high. Rehearse
realizes this finding as a Propose--Predict--Execute skill that compares candidates
and judges them from focused records of similar attempts and outcomes. Across
$4{,}000$ budgeted training runs on three loops, it improves the endpoint under
the same training-run budget. The evidence covers measurable-outcome loops with
finite candidate pools; longer runs and other pool sizes remain untested.

\setlength{\bibsep}{2pt plus 0.3pt}
\bibliographystyle{iclr2026_conference}
\bibliography{references}

\appendix
\section{Propose--Predict--Execute algorithm}
\label{sec:appendix}

\begin{algorithm}[ht]
\caption{The Propose--Predict--Execute loop. Rehearse adds the \textsc{Predict}
step and the store $\mathcal{M}$; with both removed it is a standard
autoresearch loop.}
\label{alg:ppe}
\begin{algorithmic}[1]
\State initialize store $\mathcal{M} \gets \varnothing$ and accepted baseline state $s_0$
\While{compute budget remains at step $t$}
  \State $C \gets \Call{Propose}{s_t}$ \Comment{several candidate modifications}
  \State $c^\star \gets \Call{Predict}{C, s_t, \mathcal{M}}$ \Comment{top-ranked by strict consensus}
  \State apply $c^\star$ to a copy of $s_t$ and run training, yielding result $r$
  \State $\mathcal{M} \gets \mathcal{M} \cup \{\Call{Record}{c^\star, r}\}$
  \If{$r$ improves the metric}
    \State commit the patch and set $s_{t+1}$ to the improved state
  \Else
    \State discard the patch and keep $s_{t+1}=s_t$
  \EndIf
\EndWhile
\end{algorithmic}
\end{algorithm}

All live experiments execute the single top-ranked candidate before the next
proposal round. The $K>1$ setting in Appendix~\ref{sec:appendix-precision} is
only a budgeted-selection calibration and is not used by the controller evaluated
here.

\section{Reproducibility details}
\label{sec:appendix-repro}

\paragraph{Models.} Single-turn numbers in Section~\ref{sec:single} are the
mean over three judge models---\texttt{hy3-preview}, \texttt{glm-5.1}, and
\texttt{deepseek-v4}---each run under the same protocol. In the multi-turn loops
(Section~\ref{sec:multi}) the same model drives both proposal and judging within a
loop. The live evaluation spans three serving models, one per loop---\texttt{hy3-preview}
on nanochat, \texttt{glm-5.1} on CIFAR-10, and \texttt{deepseek-v4} on ETTh1.
Because model and task are paired rather than crossed, the experiment does not
separately identify model and task effects. Model calls are issued
through the CodeBuddy framework~\citep{tencentcodebuddy2026}.

\begin{table}[H]
\centering
\small
\setlength{\tabcolsep}{4pt}
\caption{Model endpoints used in the experiments and their closest public
documentation. Endpoint IDs are CodeBuddy serving names or aliases rather than
frozen checkpoints; the citation column does not assert checkpoint identity.}
\label{tab:model-endpoints}
\begin{tabular}{@{}>{\raggedright\arraybackslash}p{0.20\linewidth}
                >{\raggedright\arraybackslash}p{0.28\linewidth}
                >{\raggedright\arraybackslash}p{0.25\linewidth}
                >{\raggedright\arraybackslash}p{0.17\linewidth}@{}}
\toprule
Endpoint & Public documentation & Experiments & Role \\
\midrule
\texttt{hy3-preview} & Exact model card~\citep{tencent2026hy3preview} & nanochat; single-turn & loop; judge \\
\texttt{glm-5.1} & Exact API documentation~\citep{zai2026glm51docs} & CIFAR-10; single-turn & loop; judge \\
\texttt{deepseek-v4} & Endpoint list and family report~\citep{deepseek2026modeldocs,deepseekai2026deepseekv4} & ETTh1; single-turn & loop; judge \\
\bottomrule
\end{tabular}
\end{table}

\paragraph{Retrieval and memory.} For each executed attempt, Rehearse derives two
views needed by the judge: the candidate description used for embedding and a
binary outcome used for the retrieved precedent. Metrics are normalized so larger
is better; a candidate is labeled \texttt{worked} iff its
post-run validation metric strictly improves over the current accepted baseline,
and crashes, timeouts, invalid patches, or non-improvements are labeled
\texttt{did not work}. Candidate descriptions are embedded with the
$384$-dimensional, $L_2$-normalized \texttt{all-MiniLM-L6-v2}
encoder~\citep{sentencetransformers2026allminilm,reimers2019sentencebert,wang2020minilm}, so dot product equals
cosine.

When a candidate is judged, the store searches only records from the same task
that were executed before the current decision. It returns every prior attempt
with cosine similarity $\ge 0.40$ to the candidate; nothing is returned below
that. The prompt surface is intentionally smaller than the stored record: each
retrieved item is serialized as one line containing only the change description
and its outcome, with no failure reason, summary, or applied-change list. The
threshold was fixed from a sweep over $\{0.30,\dots,0.60\}$
(Appendix~\ref{sec:appendix-threshold}), not tuned on the benchmark metric. The
store is persisted as plain files.

\paragraph{Selection.} In the live loops the agent proposes $N$ candidates per
step: $N=1$ for \emph{vanilla} and $N=5$ for every multi-proposal configuration,
with \emph{propose-many} evaluated only on nanochat. For \emph{select} and
\emph{Rehearse}, the predict step keeps the top $k=1$ by a pairwise
strict-consensus tournament: every unordered pair is judged in both
orders, an agreed verdict gives one vote to the winner, and a disagreement gives
no vote to either candidate. The candidate with the most votes is run; ties are
broken by the mean confidence of won comparisons and then by the fixed proposal
order. On the single-turn benchmark each item is one pair under the same
strict-consensus rule; selective accuracy is taken over pairs that receive a
verdict.

\paragraph{Benchmark construction.} We pin the public AutoSOTA source to commit
\texttt{419fe013}~\citep{autosota2026logs}; all counts and aggregates below are
our derivations from its logs, not quantities reported by AutoSOTA. The
single-turn benchmark contains $366$ pairs drawn from $39$
paper-derived AutoSOTA optimization tasks: $296$ primary
type-1 pairs, where one change worked and one did not under the same accepted
baseline, and $70$ auxiliary type-2 pairs, where both changes worked and the
larger measured gain is the answer. The type-2 set is intentionally restricted to
pairs where the earlier success is a small incremental gain, making the later
baseline close enough for an informative comparison; we report it separately
because successful attempts can still update the accepted baseline and change the
state for a later successful attempt. Each candidate is serialized to the judge
as its description, outcome-blind hypothesis, implementation, and type only; the
measured outcome, metric delta, keep/revert status, and any failure reason are
stripped before prompting, so the judge never sees an outcome. For memory
variants, retrieval is cut off before the earlier candidate in the pair is
executed, and both pair candidates are ineligible retrieval records. A
length-balanced
variant equalizes the token length of the two candidates in a pair to rule out a
length shortcut; all main numbers use it.

\paragraph{Compute.} Each experiment in a live loop is one budgeted training run under
a fixed $30$-minute wall-clock budget. Each loop runs for $100$ experiments. The
multi-turn evaluation totals $4{,}000$ budgeted training runs: $2{,}000$ on nanochat
($4$ configurations $\times$ $5$ seeds $\times$ $100$ experiments) and $2{,}000$
on the two additional loops ($2$ loops $\times$ $2$ configurations $\times$ $5$
seeds $\times$ $100$ experiments). This budget counts training runs rather than
total wall-clock or inference cost. For \emph{select}/\emph{Rehearse}, each step
adds one proposal call and $2\binom{5}{2}=20$ judge calls for the two-order
strict-consensus tournament.
\section{Prompts}
\label{sec:appendix-prompts}

\paragraph{Idea proposal.} At each step the agent is asked to read the current
training script and results log and emit a JSON array of $N$ candidates, each of
the form:
\begin{quote}\small\ttfamily
\{"description": "...", "hypothesis": "mechanistic reason this improves the
metric", "implementation": "exact change to the target training script", "type": "PARAM|CODE",
"priority": "HIGH|MEDIUM|LOW", "risk": "HIGH|MEDIUM|LOW"\}
\end{quote}
The prompt instructs the $N$ ideas to target different axes (optimizer, architecture,
regularization, schedule, data) rather than minor variants of one hyperparameter,
and states the task, metric, and compute constraints.

\paragraph{Pairwise judge.} The judge is shown the shared context, the two
candidates, and, for Rehearse, each candidate's retrieved similar prior attempts,
and returns a JSON verdict. The system prompt is:
\begin{quote}\small
You are an ML experiment advisor. Given two proposed modifications to the same
codebase targeting the same metric, predict which one is more likely to produce a
larger improvement on the target metric when executed. If
\texttt{previously\_tried\_similar} is present on an idea, it lists earlier
modifications in this task that resemble the candidate, each marked worked or did
not work (related earlier attempts, not the candidate itself). Respond with JSON
\texttt{\{"winner": "A"|"B", "confidence":
0.5--1.0, "reasoning": "..."\}}.
\end{quote}
The dump, summary, Reflexion, and reason variants of Section~\ref{sec:single-cliff}
differ only in what history block is attached to each candidate; the instruction and
output format are otherwise identical.

\section{Embedding threshold and retrieval sensitivity}
\label{sec:appendix-threshold}

The similarity threshold $0.40$ was chosen before measuring benchmark accuracy, from
a sweep over $\{0.30,0.35,0.40,0.45,0.50,0.55,0.60\}$ reporting how many pairs
retrieve at least one neighbor and how many neighbors each pair draws. Below $0.40$
retrieval pulls in loosely related attempts (more neighbors, lower relevance); above
it coverage thins. Cliff-bucket selective accuracy is flat across $0.40$--$0.55$
(within the bucket's run-to-run band), degrading only when retrieval is too loose
($0.35$) or capped to a single item. Performance is therefore insensitive to the
exact threshold within this range. We use a
small sentence embedder (\texttt{all-MiniLM-L6-v2}, 384-dim) deliberately: the store
should be cheap enough to consult every step. The reported gains are obtained
with this modest encoder; stronger retrievers remain an avenue for future work.

\section{Why focused memory works: component analysis}
\label{sec:appendix-why}

Table~\ref{tab:ablation} tests focused retrieval one ingredient at a time and
identifies three effects.
\textbf{(i) Relevant text carries the signal.} Replacing focused retrieval with
the full history dump---the same outcomes plus unrelated ones---drops cliff
selective accuracy from $83.5\%$ to $70.8\%$. Compressing the retrieved text to a
similar-count has the same effect. The summary baseline shows a related loss:
compression can omit the precedent needed for the current decision
(\S\ref{sec:single-cliff}).
\textbf{(ii) Records from worked and did-not-work attempts provide complementary evidence.} Keeping
only worked precedents lowers selective accuracy to $78.9\%$ and raises coverage
from $77\%$ to $82\%$; keeping only did-not-work precedents lowers selective accuracy
to $76.1\%$. Mixed outcomes help the judge distinguish a promising extension
from a repeated exhausted direction.
\textbf{(iii) Outcome labels carry decision signal.} Removing the worked/did-not-work
label while keeping the idea text lowers selective accuracy to $77.5\%$. The text
identifies what was tried, while the label indicates how a similar attempt
turned out. These ablations support focused, outcome-labeled
retrieval as the source of the gain.

\section{Heuristic (no-LLM) baseline}
\label{sec:appendix-heuristic}

To check that the effect is not simply having memory of unsuccessful attempts or
a similarity rule, we evaluate one illustrative rule-based selector with no LLM
judgment: for each pair, embed both candidates, measure each one's maximum cosine
similarity to the task's prior \emph{did-not-work} attempts, and pick the candidate
\emph{farther} from those attempts (abstaining when neither side has a
did-not-work precedent). This is
deduplication and direction-blacklist logic in an embedding form. On the deep
bucket it reaches $47.8\%$ ($n{=}23$ decided) and $52.4\%$ overall
($n{=}145$)---below chance on the deep bucket and near chance overall, and well
under Rehearse's $83.5\%$ selective
accuracy and even the memoryless judge's $56.9\%$. The rule is one instance rather than a proof about all
heuristics, but it illustrates a structural failure mode (Section~\ref{sec:multi-case}):
the right move is frequently a near neighbor of an attempt that did not work---raising a
hyperparameter whose lowering just did not work---so a rule that steers away from
such neighbors discards the correction together with the repeat. Distance from
unsuccessful attempts is not a proxy for whether a change will work; reading the outcomes is what the
judge does that the rule cannot.

\section{Per-seed generalization results}
\label{sec:appendix-perseed}

\begin{table}[ht]
\centering
\small
\setlength{\tabcolsep}{3.5pt}
\caption{Per-seed final improvement over baseline at experiment 100 on the two
additional loops. Rehearse ends ahead on $9$ of $10$ runs.}
\label{tab:general}
\begin{tabular*}{\linewidth}{@{\extracolsep{\fill}}ll
*{6}{S[table-format=2.1]}@{}}
\toprule
Loop & Method & {Seed 1} & {Seed 2} & {Seed 3} & {Seed 4} & {Seed 5} & {Mean} \\
\midrule
\multirow{2}{*}{CIFAR-10 (acc.\ gain \%)}
 & Vanilla  & 1.7 & 2.4 & 2.2 & 2.4 & 1.7 & 2.1 \\
 & Rehearse & \bfseries 3.6 & \bfseries 2.5 & \bfseries 2.7 & \bfseries 2.6 & \bfseries 2.8 & \bfseries 2.9 \\
\midrule
\multirow{2}{*}{ETTh1 (MSE reduction \%)}
 & Vanilla  & 31.1 & \bfseries 59.9 & 31.4 & 33.0 & 45.1 & 40.1 \\
 & Rehearse & \bfseries 52.4 & 50.2 & \bfseries 55.0 & \bfseries 59.2 & \bfseries 53.1 & \bfseries 54.0 \\
\bottomrule
\end{tabular*}
\end{table}

\section{Diminishing returns with iteration depth}
\label{sec:appendix-decay}

\begin{figure}[H]
\centering
\includegraphics[width=0.95\linewidth]{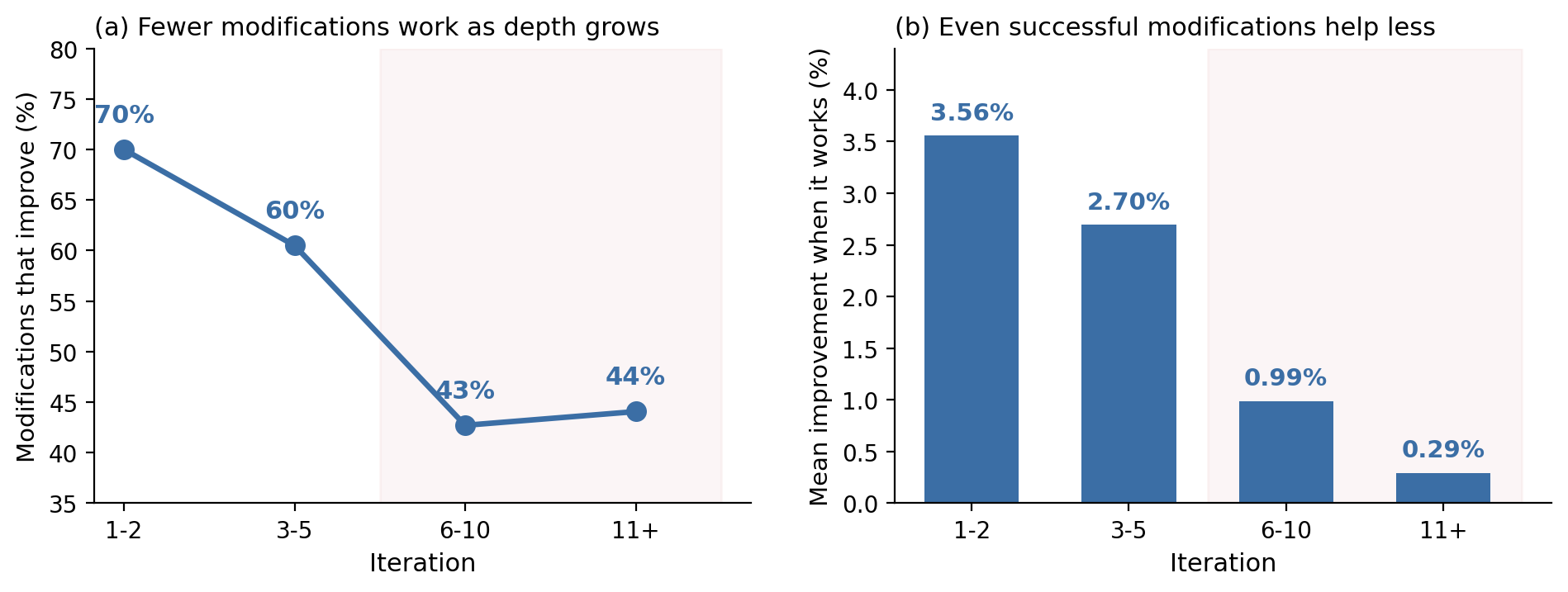}
\caption{\textbf{Late-stage returns diminish in the AutoSOTA logs.} On the AutoSOTA
optimization logs: (a)~the share of agent-proposed modifications that improve the
metric falls from 70\% in iterations 1--2 to 43\% by iteration 6+; (b)~even among
the modifications that \emph{do} work, the mean improvement shrinks from 3.6\% to
0.3\%. This motivates measuring pre-execution judgment: late in a run, most
experiments are executed and then reverted.}
\label{fig:figure0}
\end{figure}

\section{From pairwise judgments to budgeted selection}
\label{sec:appendix-precision}

The benchmark in Section~\ref{sec:single} is pairwise, but deployment uses a
pool: propose several candidates and run only a few. This appendix makes the
translation explicit for the calibration mentioned in
Section~\ref{sec:single-selection}. The live-loop setting in
Section~\ref{sec:multi} uses $N=5$ and $K=1$; the 10-candidate, keep-three table
below is an illustrative budgeted-selection setting, not the live-loop protocol.

Assume a pool of $N$ candidates, of which $W=rN$ actually work, and a budget
of $K$ executed candidates. Precision@K is
\[
  \mathrm{Precision@}K = \frac{1}{K}\sum_{i \in S_K} y_i,
\]
where $S_K$ is the selected set and $y_i \in \{0,1\}$ indicates whether candidate
$i$ worked. For uniformly random selection,
\[
  \mathbb{E}\!\left[\#\{i \in S_K: y_i=1\}\right] = K\frac{W}{N},
  \qquad
  \mathbb{E}\!\left[\mathrm{Precision@}K\right] = \frac{W}{N}=r.
\]
Thus the random Precision@K baseline is exactly the pool worked rate. An oracle
that knows the outcomes has
\[
  \mathrm{Precision@}K_{\mathrm{oracle}} = \min\!\left(1,\frac{W}{K}\right).
\]

For pairwise selection, we use a coverage-aware approximation to the
strict-consensus tournament. Let $c$ be coverage, the probability that the two
presentation orders agree, and let $q$ be selective accuracy conditioned on that
agreement. When a worked candidate is compared with a candidate that did not work,
\[
  P(\text{correct vote})=cq,\qquad
  P(\text{wrong vote})=c(1-q),\qquad
  P(\text{no vote})=1-c.
\]
For same-outcome comparisons, a vote is produced with probability $c$ and
assigned uniformly; otherwise neither candidate receives a vote. The tournament
score is the number of votes, and the top $K$ candidates are selected with random
tie-breaking.

For this conversion, No memory uses $(q,c)=(0.795,0.824)$ and Rehearse uses
$(q,c)=(0.842,0.833)$. The tables report the resulting Monte Carlo expectations
under independent pairwise events and random score tie-breaking while preserving
strict-consensus no-votes.

\begin{table}[H]
\centering
\small
\setlength{\tabcolsep}{5pt}
\caption{\textbf{Live-sized selection calibration.} Expected success@1 (\%)
for selecting $K=1$ candidate from a pool of $N=5$, using the coverage-aware
tournament model above.}
\label{tab:precision-live}
\begin{tabular}{@{}S[table-format=2.0]
*{4}{S[table-format=3.1]}@{}}
\toprule
{Worked (\%)} & {Random} & {No memory} & {Rehearse} & {Oracle} \\
\midrule
20 & 20.0 & 54.3 & 61.2 & 100.0 \\
40 & 40.0 & 79.7 & 85.4 & 100.0 \\
60 & 60.0 & 91.7 & 94.8 & 100.0 \\
80 & 80.0 & 97.3 & 98.5 & 100.0 \\
\bottomrule
\end{tabular}
\end{table}

\begin{table}[H]
\centering
\small
\setlength{\tabcolsep}{4.5pt}
\caption{\textbf{Illustrative budgeted-selection calibration.} Expected
Precision@3 (\%) for selecting $K=3$ candidates from a pool of $N=10$.}
\label{tab:precision}
\begin{tabular}{@{}S[table-format=2.0]
*{4}{S[table-format=3.1]}@{}}
\toprule
{Worked (\%)} & {Random} & {No memory} & {Rehearse} & {Oracle} \\
\midrule
10 & 10.0 & 27.6 & 29.7 & 33.3 \\
20 & 20.0 & 51.9 & 56.4 & 66.7 \\
30 & 30.0 & 71.1 & 77.5 & 100.0 \\
50 & 50.0 & 90.6 & 94.6 & 100.0 \\
70 & 70.0 & 97.3 & 98.8 & 100.0 \\
90 & 90.0 & 99.5 & 99.8 & 100.0 \\
\bottomrule
\end{tabular}
\end{table}

\end{document}